\documentclass[fleqn,10pt]{wlscirep}
\usepackage[utf8]{inputenc}
\usepackage[T1]{fontenc}
\usepackage[font=small]{subcaption}
\usepackage[font=small,labelfont=bf]{caption}
\usepackage{gensymb}
\usepackage{url}

\title{Forecasting large-scale circulation regimes using deformable convolutional neural networks and global spatiotemporal climate data}

\author[1,2,*]{Andreas Holm Nielsen}
\author[1]{Alexandros Iosifidis}
\author[1]{Henrik Karstoft}  

\affil[1]{Aarhus University, Department of Electrical and Computer Engineering, Aarhus, Denmark}
\affil[2]{Danske Commodities A/S, Aarhus, Denmark}

\affil[*]{ahn@ece.au.dk}

\begin{abstract}

Classifying the state of the atmosphere into a finite number of large-scale circulation regimes is a popular way of investigating teleconnections, the predictability of severe weather events, and climate change. Here, we investigate a supervised machine learning approach based on deformable convolutional neural networks (deCNNs) and transfer learning to forecast the North Atlantic-European weather regimes during extended boreal winter for 1 to 15 days into the future. We apply state-of-the-art interpretation techniques from the machine learning literature to attribute particular regions of interest or potential teleconnections relevant for any given weather cluster prediction or regime transition. We demonstrate superior forecasting performance relative to several classical meteorological benchmarks, as well as logistic regression and random forests. Due to its wider field of view, we also observe deCNN achieving considerably better performance than regular convolutional neural networks at lead times beyond 5-6 days. Finally, we find transfer learning to be of paramount importance, similar to previous data-driven atmospheric forecasting studies.

\end{abstract}
\begin{document}

\flushbottom
\maketitle
\thispagestyle{empty}

\section*{Introduction}\label{intro}

In the dynamical meteorology literature, it is common practice to represent the state of the atmosphere with a finite number of weather regimes or large-scale circulation patterns \cite{michelangeli1995, Cassou2008}. This has several useful purposes, such as the investigation of global teleconnections \cite{Cassou2008, lee2019enso}, analyzing predictive skill of numerical weather prediction (NWP) models \cite{Ferranti2018, Albers_2021}, and quantifying climate change \cite{deser2017role}. For example, in the North Atlantic-European (NAE) region, the ability to accurately predict large-scale circulation regimes is directly linked with the predictability of severe temperature anomalies such as the sequences of cold spells over northern and western Europe in the winter of 2009/2010 \cite{Ferranti2018}. Interestingly, due to complex climate interactions across numerous spatial and temporal scales, multiple studies \cite{steinhaeuser2012multivariate, casagrande2015wavelet, paluvs2018linked, agarwal2019network} suggest we have an insufficient understanding of global climate processes. As a result, improving our understanding and skill in forecasting weather regimes and teleconnections provides considerable opportunities for an improved understanding of the global climate system and increased predictability of the atmosphere at lead times beyond the typical two-week limit stated in the literature \cite{Rasp2020, Kawale2013}.

Machine learning (ML), and deep learning (DL) in particular, has recently achieved substantial success in multiple fields of science \cite{Rasp2020}. This success is largely attributed to deep learning models being able to effectively learn patterns in high-dimensional data rather than being explicitly formulated by domain experts. This has enabled advances in, for example, weather forecasting \cite{metnet2, Rasp2020}, parameterization of convection \cite{gentine2018could}, and detection of tropical cyclones and synoptic-scale fronts \cite{kunkel2018automated, liu2016application}. 

The application of deep learning specifically for identifying and predicting weather regimes, climate, and teleconnections has also recently started to attract interest \cite{ensodeeplearn, chatto}. However, previous data-driven studies are typically limited to a) forecasting circulation regimes at time horizons such as five days ahead, where NWP models are already considered accurate \cite{Ferranti2018}, or b) not having the ability to model and incorporate global weather data and teleconnections \cite{ensodeeplearn}. Furthermore, while recent studies have achieved success applying data-driven regularized regression methods for climate prediction \cite{delsole2017statistical, stevens2021graph}, these do not provide as information-rich output like related probabilistic methods \cite{ensodeeplearn}.    

As a result, the investigation of probabilistic data-driven methods for predicting large circulation regimes at the extended medium-range time horizon (days to weeks ahead) with global weather data remains largely underexplored. This is despite major circulation regimes being a primary driver of weather fluctuations at this time horizon \cite{lee2019enso}, while also being the breaking point where various global NWP models tend to show little to no skill in predicting these regimes \cite{Ferranti2018}.

This paper investigates data-driven probabilistic forecasting of NAE weather regimes \cite{michelangeli1995, Cassou2008} for 1 to 15 days ahead using 182 years of daily multivariate weather data. To achieve our stated goal, we use deformable convolutional neural networks (deCNN), a modified version of regular convolutional neural networks (CNNs) specifically tailored to overcome a key limitation of regular CNNs, namely learning transformation-invariant features with much greater localization capability and across global regions \cite{Dai2017}. To validate the performance of the deCNN model, we compare it with a regular CNN and several meteorological and machine learning benchmarks using multiple performance metrics. In addition, we maintain a high emphasis on interpretability by using sophisticated interpretation techniques from the machine learning community. This enables us to attribute the contribution of each weather variable for a given observation on a grid point basis. This focus on explainable machine learning has been paramount to data-driven research in the atmospheric sciences for the past few years \cite{McGovern2019}. In addition, recent studies have even argued that the interpretation of neural networks might be more important than the model output itself for enabling scientific discovery within the geosciences \cite{toms2020physically}. Taking weather regimes as an example, we could potentially discover new teleconnections targeted towards specific weather regimes or regime transitions by using these interpretation techniques. We expect this to be incredibly useful in its own right, given the rife amount of existing teleconnections and the current difficulty of attributing particular teleconnections towards specific regions or climates.

\section*{Data and Methods}

To maximize the benefit of deep learning and data-driven methods, one typically requires a large body of (preferably) labeled data to achieve the best results \cite{chatto}. To tackle this challenge, we use the popular method of labeling NAE weather regimes using empirical-orthogonal functions (EOF) and the k-means algorithm \cite{michelangeli1995, Cassou2008}, which results in four well-established regimes or classes. These are computed on both a large reanalysis dataset and a shorter and more up-to-date model reanalysis, effectively obtaining 182 years of daily weather regimes. To explain and predict these regimes 1 to 15 days ahead, we use geopotential height anomalies at 50, 200, 500, and 700 hPa, in addition to sea surface temperature (SST) anomalies and mean sea level pressure (MSLP) anomalies. These are some of the most used variables for defining and predicting teleconnections \cite{Cassou2008, Bloomfield2020, Ferranti2018, lee2019enso}. In addition, we use the popular transfer learning technique from the machine learning community to account for systematic differences between our two datasets while also reducing overfitting and improving forecast skill \cite{rasp2021data}. This method works by pretraining our model on the large dataset and transferring the knowledge learned to the smaller dataset. More specifically, the model pre-trained on the large dataset will function as a sophisticated initialization of all parameters present for our deCNN model and the benchmark CNN model. When retraining the respective model for the smaller reanalysis dataset, all parameters are then adjusted or finetuned. This has the benefit of being more sample-efficient for the considerably smaller but more recent reanalysis dataset. This section will describe these two datasets in more detail, explain the labeling technique more thoroughly, and finally describe the methodological setup.

\subsection*{Data description and preparation}
The large dataset used in this study is based on the recent NOAA-CIRES-DOE Twentieth Century Reanalysis v3 (20CRv3) dataset \cite{slivinski2019towards}, where we have samples ranging from 1836 to 1980 with numerous weather parameters. The resolution of the dataset is approximately 75km at the equator, with 64 vertical levels and 80 individual ensemble members. We use the ensemble mean unless otherwise noted. Samples are also available from 1981 to 2015 with a different assimilation scheme, but this would overlap with our smaller transfer learning dataset to be described shortly, and as a result, we leave these samples out. Once we have trained our model on the 20CRv3 reanalysis dataset, we retrain our model on the ERA5 reanalysis dataset \cite{era5} ranging from 1981 to 2018. ERA5 has a higher resolution than 20CRv3, so to ensure transfer learning is done properly, we use a grid resolution of 5.625 degrees for both datasets and regrid all variables using bilinear interpolation. This matches the resolution used in the seminal WeatherBench paper \cite{Rasp2020} and provides gridded global samples of size $32 \times 64$ grid points per variable. Anomaly maps are then derived by subtracting daily raw samples from a 30-year monthly climatology updated and moved every five years \cite{world2017wmo}. More specifically, the first monthly climatology will cover the period 1836-1866. The second monthly climatology will then be moved five years to cover the period 1841-1871, which is then used to derive anomaly maps for the next five years extending from the previous anomaly period, i.e., 1867-1871. This continues until anomaly maps have been derived for all years in the dataset. This ensures we, at least partly, account for the influence of a changing and dynamic climate. Subtracting a monthly climatology also ensures we remove the seasonal cycle and hence the influence of autocorrelation to some extent. For more specific details on the climatology procedure, we kindly refer to guidelines provided by the World Meteorological Organization \cite{world2017wmo}. Finally, anomaly maps are used in the labeling scheme that we present shortly in more detail. In total, our datasets have 21,931 samples for 20CRv3 and 6,050 for ERA5, consisting of extended boreal winter days (15 November to 31 March)\cite{michelangeli1995}. We observed a few days with missing or corrupt values arising due to the regridding procedure, and these were removed from the dataset. This covers our explanation of the explanatory variables, and we now turn to our target variable in more detail.

\subsection*{Labeling of NAE regimes}\label{labeling}
For labeling the NAE weather regimes used for classification purposes, we follow the approach outlined by Michelangeli \textit{et~al.} \cite{michelangeli1995} and Cassou  \cite{Cassou2008}, giving us four winter NAE regimes called the positive North Atlantic Oscillation (NAO+), NAO-, Scandinavian Blocking (SB), and Atlantic Ridge (AR). These regimes are occasionally called ``Michelangeli-Cassou'' patterns (MCPs) \cite{Bloomfield2020}. The labeling is done as follows. First, we derive daily geopotential height anomalies at 500 hPa (Z500) for extended boreal winter over the domain 90\degree W--30\degree E, 20\degree--80\degree N. Then, we perform an EOF analysis, where we project the resulting anomaly maps into a 14-dimensional EOF space, corresponding to around 90\% of the total variance explained \cite{Cassou2008}. Finally, we use k-means clustering on the 14-EOF time series to extract four centroids, considered the optimal number of centroids \cite{michelangeli1995, Cassou2008}, and assign each observation to the nearest cluster in the 14-dimensional EOF space measured by minimum Euclidean distance, corresponding to the four previously mentioned NAE regime labels. These labels constitute what our deep learning is going to forecast. The most frequent weather pattern is NAO+ (32\%) followed by SB (28\%), and the least frequent ones are NAO- (19\%), followed by AR (21\%) \cite{Bloomfield2020}. We note this slight imbalance might constitute an issue when training our neural network, as deep learning models are not known to handle class imbalance well unless explicitly handled in the training process either via sampling techniques or a modified loss function \cite{lin2017focal}.  We discuss this further under the training details.

\subsection*{Deformable Convolutional Neural Networks}
As stated in the introduction, we apply deformable convolutional neural networks (deCNNs) in this study, a modified and improved version of regular CNNs \cite{Dai2017}. CNNs are limited to fixed geometric transformations, whereas deCNNs can learn transformation-invariant features with much greater localization capability and across larger regions \cite{Dai2017}. This is especially relevant in our application, where we use global gridded data and try to directly incorporate and model existing global teleconnections with the potential of uncovering new ones. The importance of having a wider field-of-view was found critical in Espeholt \textit{et~al.}\cite{metnet2} using dilated convolutions, but deformable convolutions are actually a generalization of dilated convolutions \cite{Dai2017}. 

The basic architecture of our model consists of deCNN layers followed by batch normalization, leaky rectified activation unit (leaky ReLU), and dropout. A standard CNN is a regularized version of a regular feedforward neural network, where we assume the input data is 2D or 3D and use so-called convolutional filters or kernels to extract spatial information in a more meaningful and efficient way. The output of each convolutional filter is called a feature map, and the number of filters for any given convolutional layer will form the output of that layer. In this regard, deCNNs and CNNs are the same. DeCNNs, on the other hand, augment the regular $N \times N$ kernel matrix using learnable offsets \cite{Dai2017}. This implies that deCNN operates on a potentially irregular grid but with the same spatial resolution as CNNs. This explains the greater localization capability of deCNNs and their wider field of view. When training deCNNs, the convolutional kernels for generating feature maps and learning the offsets are trained simultaneously using backpropagation. For more details regarding the differences between CNNs and deCNNs, we kindly refer to the original deCNN paper \cite{Dai2017}. The next layer, batch normalization, simply standardizes the outputs of any given layer by subtracting and dividing by the mean and variance computed over any given mini-batch, respectively \cite{batchnorm}. This typically stabilizes and speeds up training for deep neural networks \cite{batchnorm}. Dropout is a simple and effective regularization technique that randomly drops a pre-defined fraction of hidden units during backpropagation \cite{srivastava2014dropout}. Leaky ReLU \cite{lrelu} is a simple extension of the regular ReLU activation function, where we allow a small slope for negative values instead of clipping them. After these operations, we use max-pooling to extract the maximum value from a $2 \times 2$ kernel with stride one, thus effectively reducing the spatial dimension of the input layer by a factor of two. Finally, after propagating through all CNN layers, we flatten the resulting matrix and use two fully connected layers with dropout and leaky ReLU in between to finally perform multi-class classification.

\subsection*{Training details}
Splitting spatiotemporal data into a training, validation, and test set with a large dataset such as the one used in this study might seem trivial at first. However, deciding how to split the data is anything but trivial and can potentially alter the conclusions drawn from a data-driven study \cite{Rasp2020, xu2018splitting}. In this regard, stratified sampling is an efficient sampling strategy, where we split a population into groups that, in most cases, produce superior results compared to straightforward sampling \cite{dunn2011exploring}. However, defining such groups is not always straightforward. In our case, we know from previous research that teleconnections greatly impact weather cluster predictability. For example, it has been shown that MJO phases are precursors for persistent NAE weather regimes at around the ten days timescale \cite{Cassou2008, lee2019enso}, but these precursors are themselves modulated by other teleconnections such as ENSO \cite{lee2019enso} and the Pacific Decadal Oscillation (PDO) \cite{agarwal2019network}. Another important teleconnection for the NAE region is the Atlantic Multidecadal Oscillation (AMO) \cite{borgel2020atlantic}. Stratifying the performance of NWP models based on these different teleconnections and their modulating behavior suggests systematic differences in the predictive power for weather forecasts in the NAE region from 10 to 30 days into the future \cite{Ferranti2018, lee2019enso}. As a result, we argue that the optimal approach for splitting a dataset in an application like this is based on a stratified sampling using some of the most important decadal and multidecadal teleconnections for the NAE weather regimes.

We will now explain our data splitting strategy in more detail. First, our test set is defined as the period 2011 to 2018 using ERA5. We note that this 8-year period will not cover all possible teleconnection patterns. Nevertheless, we select our test set in this way to ensure our model generalizes to future data, but primarily to ensure the test set is completely separate from any training or validation set, for example, to avoid the influence of oceanic memory \cite{ensodeeplearn}. 
For our training and validation sets, we employ the stratified sampling technique based on the ENSO, PDO, and AMO phases \cite{lee2019enso}. First, we collect data from 1854 to 2018 for AMO \cite{amonoaa}, 1856 to 2018 for PDO \cite{pdonoaa}, and 1871-2018 for ENSO \cite{ensonoaa}. For 1837-1870, where we do not have all teleconnections available, we simply use this period as training data to avoid discarding over 30 years of data. After collecting the required data, we find the percentage of samples for all phases based on a) the 20CRv3 period (1870-1979) and b) the reduced ERA5 period without the test set (1979-2011). Then, we randomly sample 20\% into a validation set and leave the remaining 80\% as training data. Here, we preserve the previously computed distributions for all phases as much as possible for both the validation and training set. To ensure we do not have overlap between the randomly stratified samples, we restrict each subsampled fold to contain at least one year of data.

We use weighted cross-entropy loss as our loss function, where the weights are defined as the ratio of samples for any given class relative to the number of samples for the majority class, thus being a number from 0 to 1. This is a simple way of dealing with class imbalance, as we assume all classes are equally important despite differences in the number of samples for each. We also use one of the most popular gradient optimizers, the Adam optimizer \cite{adam}. To reduce overfitting, we apply the early stopping technique, which works by stopping training early once a pre-defined metric such as accuracy stops improving on the validation set. 
Finally, we use Bayesian Optimization (BO) to select the optimal hyperparameter set. BO is, in simple terms, a sample-efficient global optimization method where we direct search based on performance on the validation set from previous iterations \cite{snoek2012practical}. This technique is often much faster at finding an optimum compared to regular hyperparameter tuning approaches that do not direct search in any way, such as grid search or random search \cite{agnihotri2020exploring}. We use BO for each lead time independently, meaning we will potentially have a unique set of hyperparameters for each lead time. We could also have trained an aggregate model for all lead times with appropriate lead time encoding similar to other studies \cite{metnet2}. However, this tends to yield similar performance \cite{metnet2} and is mostly relevant when training time is a bottleneck. This is not the case in our framework, as considerable parallelization is possible, and training time is around 30 minutes per lead time, including pretraining, on two Nvidia RTX 2080-RTI GPUs. The hyperparameters we optimize are a) the number of filters for our deCNN and CNN modules, b) the dropout rate, c) the number of fully connected neurons in the penultimate layer, and d) the batch size due to its effect on regularization \cite{smith2018disciplined}. Finally, we select the optimal learning rate based on the popular ``LR range test'' as formulated by Smith \cite{smith2017cyclical}.

\subsection*{Performance Evaluation}
To evaluate our deep learning model, we implement several different probabilistic and categorical performance metrics that we evaluate at all lead times from 1 to 15 days into the future. These include Area Under the Receiver Operating Characteristic Curve (AUC-ROC), Critical Success Index (CSI), Brier Score (BS), and regular multi-class accuracy. The AUC-ROC metric is one of the most popular metrics for classification problems, as it incorporates much more information from the contingency matrix than regular classification accuracy \cite{metz1978basic}. The ROC curve itself plots the probability of detection versus the probability of false detection with various thresholds for a given model in the binary classification case, and anything above the diagonal is considered better than random chance \cite{hand2001simple}. One way to compute the AUC-ROC for multi-class classification problems is by weighting each class-specific AUC-ROC metric by the number of true instances for that class, essentially treating all classes as individual binary classification problems \cite{hand2001simple}. This has the advantage of taking class imbalance into account. In short, the AUC-ROC metric measures how well a model can differentiate between classes at different thresholds by computing the AUC-ROC metric.

The next metric, CSI, is defined as 
\begin{equation}
    CSI = TP / (TP+FN+FP)
\end{equation}\label{eq:csi}
where TP is the number of true positives, FN the number of false negatives, and FP the number of false positives. CSI, like AUC-ROC, is not directly a multi-categorical metric, and as a result, we use the same weighted scoring approach as for AUC-ROC. 

The Brier Score in the binary case is simply defined as the mean squared error of a probabilistic forecast:

\begin{equation}
BS=\frac{1}{N} \sum^N_{t=1}(f_{t} - y_{t})^{2}
\end{equation}
where $f_t$ denotes the forecast probability from 0 to 1 for a point in time $t$, $y_t$ is the ground truth label being a 0 or 1, and $N$ denotes the number of samples. Again, we use the weighted scoring approach to extend this to the multi-class setting. The final metric, multi-class classification, is simply defined as the average correspondence between individual pairs of ground truth labels versus predicted labels. Finally, we include a ROC curve and a performance diagram for our deCNN model. Performance diagrams are popular in the weather forecasting community as they summarize both probability of detection, false alarm ratio, bias, and critical success in one figure \cite{roebber2009visualizing}. The upper right corner of the figure denotes the area where performance is best, and the diagonal denotes various frequency bias levels.

Next, we discuss the benchmark models used in this study. The first two benchmark models we include are logistic regression (LR) and random forests (RF) due to their popularity and strong performance in the atmospheric machine learning community \cite{chatto, jergensen2020classifying, gagne2017storm, McGovern2019}. These use the same training and validation framework described earlier for our deep learning model. However, since RF tends to work poorly with raw images by treating all pixels as different features, it lacks the flexibility of, e.g., CNNs to perform feature processing and learn spatial structures gradually. As a result, one typically uses feature transformation techniques beforehand on the input used by the RF model \cite{bosch2007image}.

This is also consistent with the seminal paper of Ham \textit{et~al.} \cite{ensodeeplearn}, where they use EOFs as input to benchmark models for forecasting the El Niño-Southern Oscillation (ENSO). As a result, we use the 20 EOF components from Z500 as input to both the LR and RF benchmarks. Importantly, as these 20 components are orthogonal and represent over 90\% of the variance explained \cite{Cassou2008}, we avoid correlated features, which can pose a significant problem for LR \cite{multicol}, at the expense of relatively little information lost. We did investigate using raw data instead, similar to Chattopadhyay \textit{et~al.} \cite{chatto}. However, our experiments consistently showed a decline in validation and test performance for both LR and RF of around 5-10\%. This slightly contradicts the findings of Chattopadhyay \textit{et~al.} \cite{chatto}, which we assume could be due to multiple reasons, including a) our dataset spanning a much larger range from 1836 to 2018 in contrast to their 1980 to 2005, or b) the mixing or inclusion of both summer and winter circulation patterns in Chattopadhyay \textit{et~al.} \cite{chatto}, in comparison to our NAE weather regimes being winter-exclusive per definition \cite{michelangeli1995, Cassou2008}. 

In addition to these benchmarks, we include the persistence forecast and climatological forecast. Persistence is arguably the simplest meteorological benchmark possible, as it simply refers to the weather regime at initialization time, $t_{0}$, propagated into the future for any given lead time. The climatological forecast is based on a weekly climatology approach similar to Rasp \textit{et~al.} \cite{Rasp2020}, in which we compute the dominant weather regimes for each of the 52 calendar weeks on the training set and then use these as weekly forecasts for our test set. It is common practice to include these two benchmarks in atmospheric forecasting studies as they constitute the minimum requirement for a model to have forecasting skill \cite{Rasp2020}.

\section*{Results}
We present our results based on the previously defined test set in two sections. First, we present the forecasting results from all models at the 1 to 15 days lead time, and then we present the interpretation results.




\begin{table}[t]
\begin{tabular}{l||ccc|ccc|ccc|ccc}

& \multicolumn{3}{c}{\textbf{ACC}}  & \multicolumn{3}{c}{\textbf{AUC-ROC}} & \multicolumn{3}{c}{\textbf{CSI}} & \multicolumn{3}{c}{\textbf{BS}}  \\ 
\textbf{Model / Days} & 5 & 10 & 15 & 5 & 10 & 15 & 5 & 10 & 15 & 5 & 10 & 15 \\ \hline \hline

Persistence & .470 & .344 & .317 & x & x & x & x & x & x & x & x & x \\ 
Weekly Climatology & .325 & .323 & .324 & x & x & x & x & x & x & x & x & x \\ \hline
LR & .481 & .336 & .311 & .739 & .614 & .557 & .315 & .205 & .179 & .641 & .726 & .745 \\ 
RF & .477 & .361 & .321 & .747 & .627 & .559 & .313 & .215 & .175 & .630 & .713 & .754 \\ \hline
CNN & .523 & .403 & .319 & .779 & .649 & .580 & .356 & .248 & .184 & .616 & .692 & .729 \\
deCNN (no pretrain) & .459 & .324 & .257 & .746 & .596 & .564 & .292 & .182 & .113 & .663 & .731 & .787 \\ 
\textbf{deCNN} & \textbf{.533} & \textbf{.422} & \textbf{.352} & \textbf{.793} & \textbf{.678} & \textbf{.632} & \textbf{.362} & \textbf{.259} & \textbf{.201} & \textbf{.604} & \textbf{.671} & \textbf{.698} \\ 
\end{tabular}
\caption{Performance metrics for all evaluated models for the four North Atlantic-European circulation regimes for daily samples between 2011-2018 relative to ERA5. Days refer to forecast lead time. Best models are highlighted in bold. Cells with ``x'' denotes unavailable values due to not being probabilistic forecasts.}
\label{tab:scores}
\end{table}

\begin{figure}[t]
  \begin{subfigure}{0.5\textwidth}
    \includegraphics[width=\linewidth]{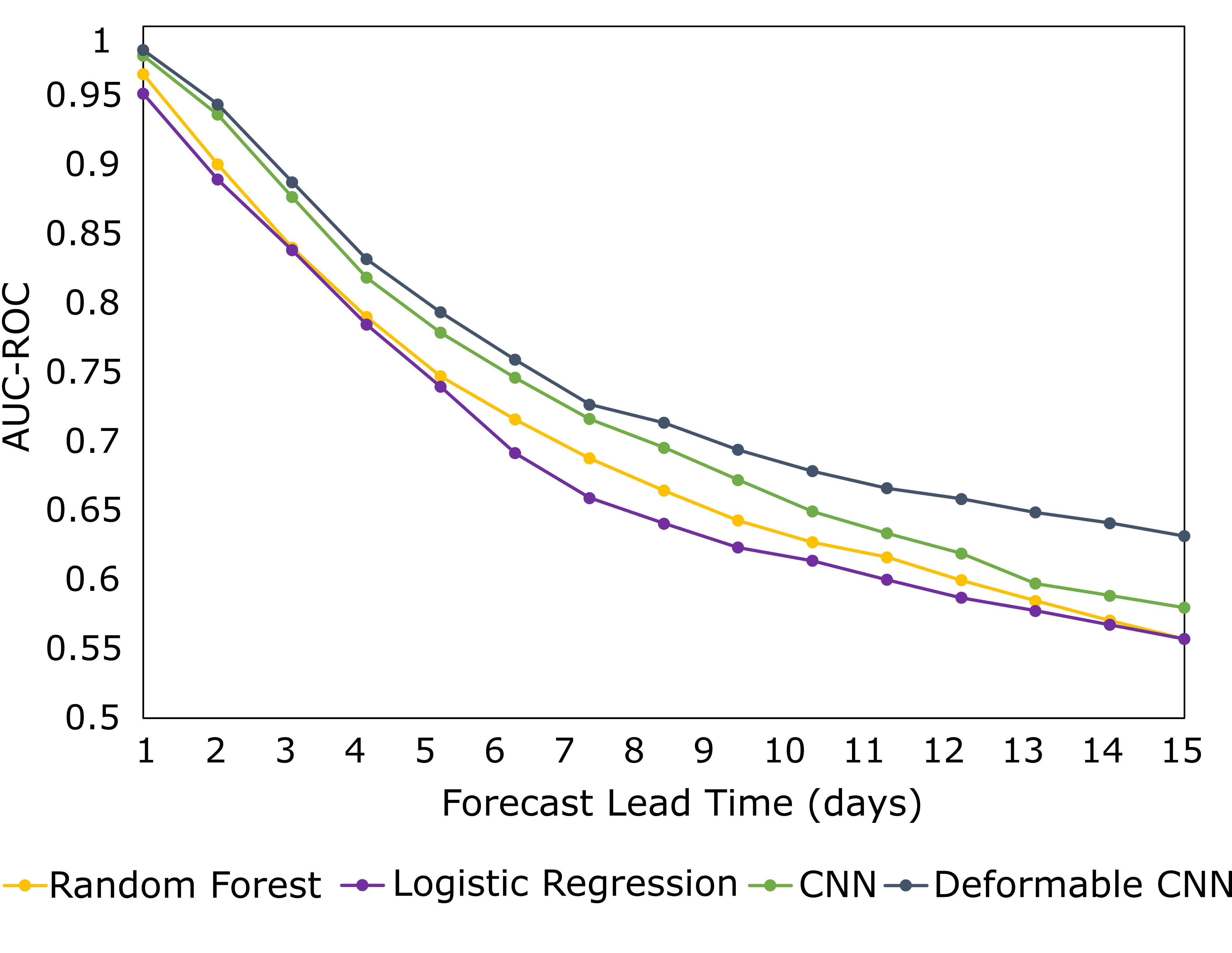}
    \caption{Area Under the Receiver Operating Characteristics Curve (AUC-ROC)} \label{fig:auc_acc_a}
  \end{subfigure}%
  \hspace*{\fill}   
  \begin{subfigure}{0.5\textwidth}
    \vspace{-4mm}
    \includegraphics[width=\linewidth]{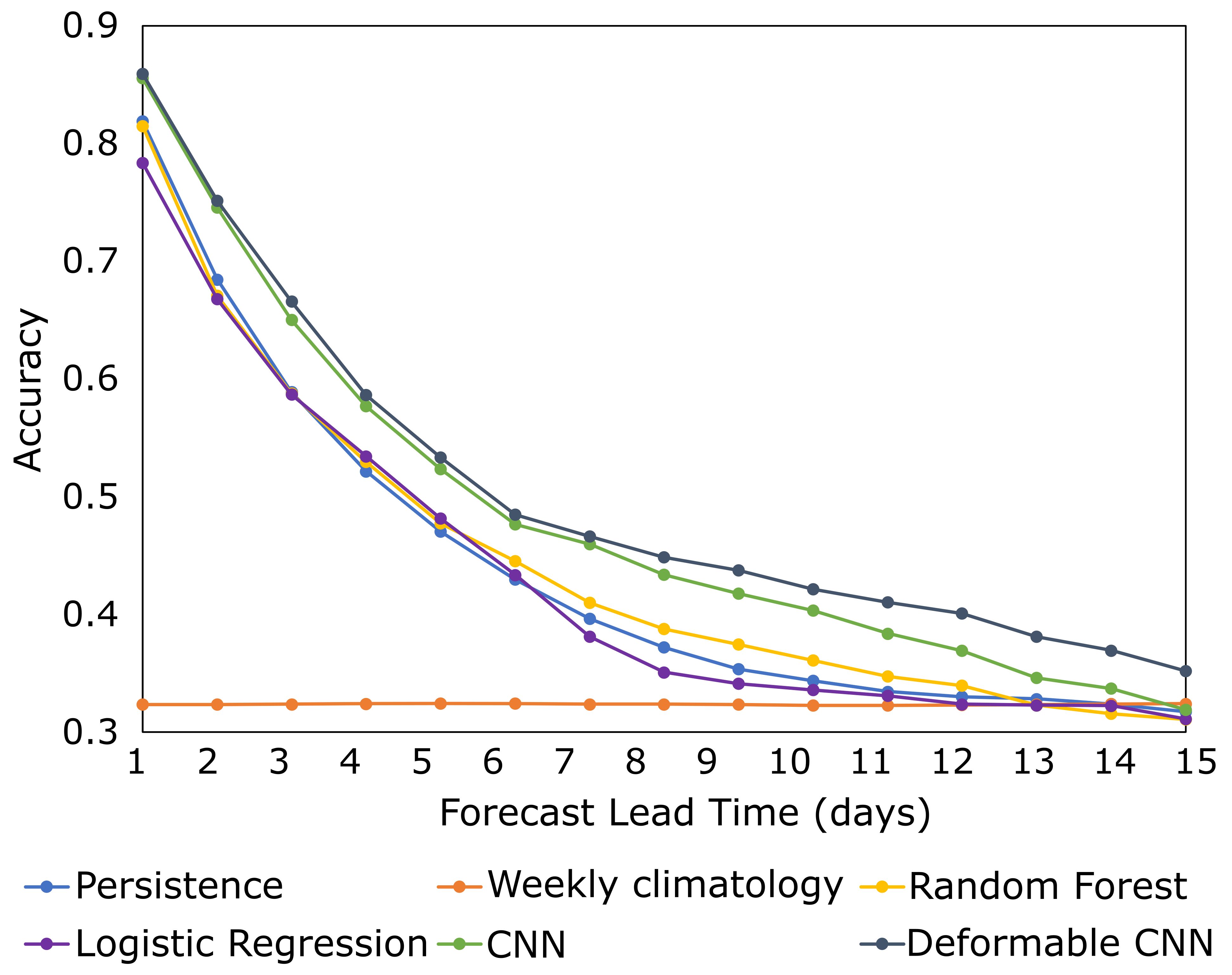}
    \caption{Accuracy} \label{fig:auc_acc_b}
  \end{subfigure}
\caption{AUC-ROC and accuracy metrics for all lead times for our proposed models. AUC-ROC is not supplied for persistence and climatology given their lack of probabilistic forecasting.} \label{fig:auc_acc}
\end{figure}

\subsection*{Forecasting results}
The results from our performance evaluation framework are summarized in Table \ref{tab:scores}. First, we observe that deCNN achieves the highest accuracy at the 5, 10, and 15 days lead times. Second, we observe that the relative improvement of deCNN over CNN gets increasingly higher at the 10 days lead time and even higher at 15 days, demonstrating the benefit and importance of having a wider field of view at longer lead times. This is also evident when looking at the similar performance between CNN and LR at the 15 days lead time.

We want to highlight that classification accuracy is not our main variable of interest, given that we want to forecast all regimes equally well from a probabilistic point of view. However, since our weather regimes are deterministic by nature, persistence and weekly climatology are also deterministic. As a result, we still include accuracy, but we note that the main performance metric of interest is the AUC-ROC despite being unavailable for the meteorological benchmarks. Nevertheless, the findings are largely unchanged. One difference can be noted at the 15 days lead time. Here, the percentage point difference between CNN and RF is only 0.08 for accuracy but 0.23 for AUC-ROC. The same pattern can be observed for deCNN relative to LR and RF. The reason here is undoubtedly the usage of weighted cross-entropy as our loss function, which might trade off having a higher AUC-ROC at the cost of a lower accuracy. We also want to highlight the strong performance of persistence and weekly climatology similar to previous studies \cite{Kumler2018, Rasp2020}. If we investigate LR and RF in more detail, we notice they are surprisingly close to both persistence and climatology at all lead times. However, at the 15 day lead time, LR and RF even show inferior performance compared to weekly climatology.

Finally, we show the importance of transfer learning for the deCNN model (no pretrain). The no pretrain deCNN model is exclusively trained using the ERA5 dataset, discarding the period 1836 to 1980 for the 20CRv3 dataset. The decline in performance by not using transfer learning is substantial, going from 0.533 to 0.459 at the 5 days lead time and from 0.352 to 0.257 at the 15 days lead time. Thus, similar to related studies \cite{ensodeeplearn, rasp2021data}, we find transfer learning critical to success in our atmospheric forecasting problem.

\begin{figure}[!htb]
  \begin{subfigure}{0.5\textwidth}
    \includegraphics[width=\linewidth]{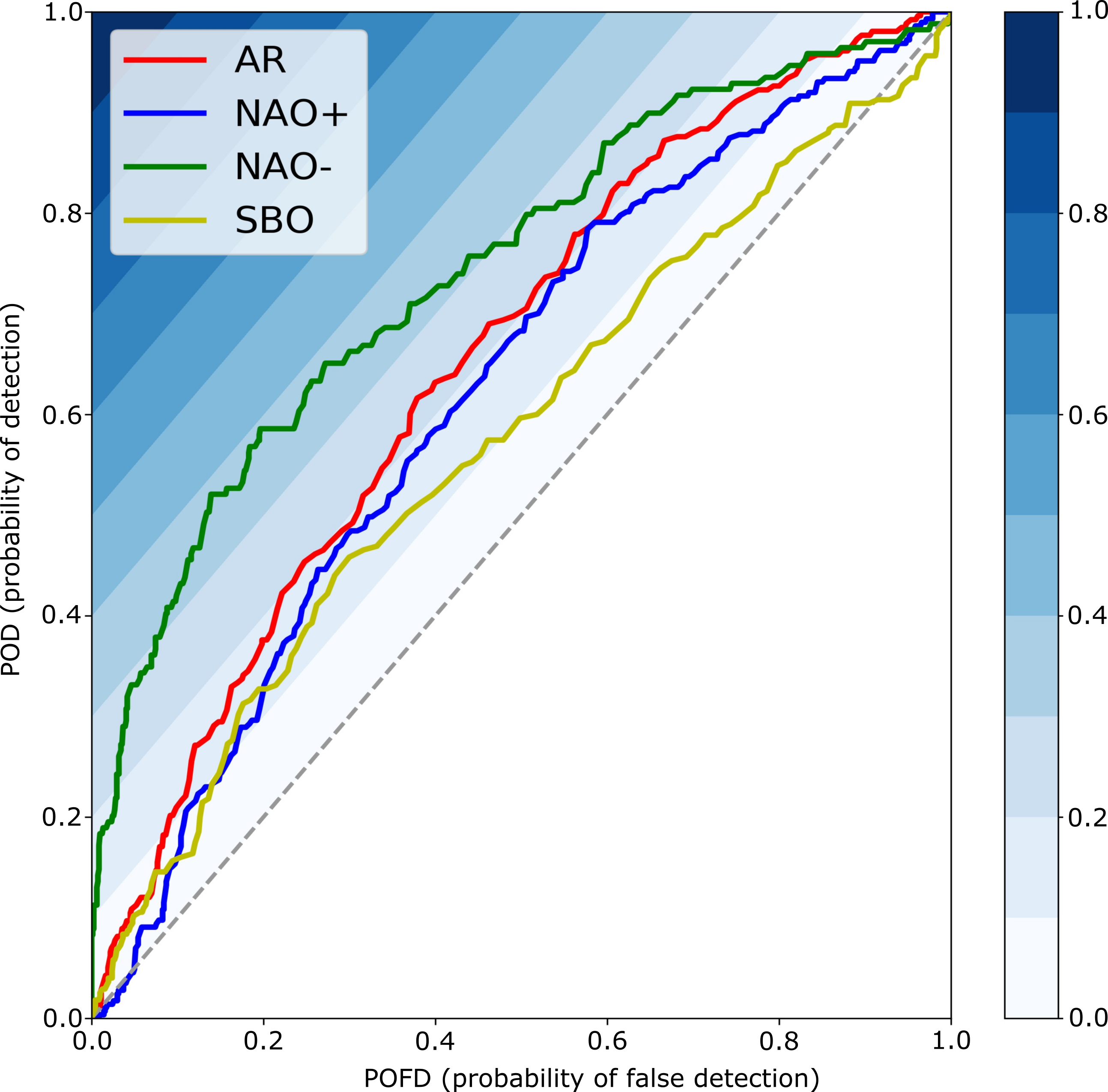}
    \caption{ROC curve} \label{fig:roc_perf_di_a}
  \end{subfigure}%
  \hspace*{\fill}   
  \begin{subfigure}{0.5\textwidth}
    \includegraphics[width=\linewidth]{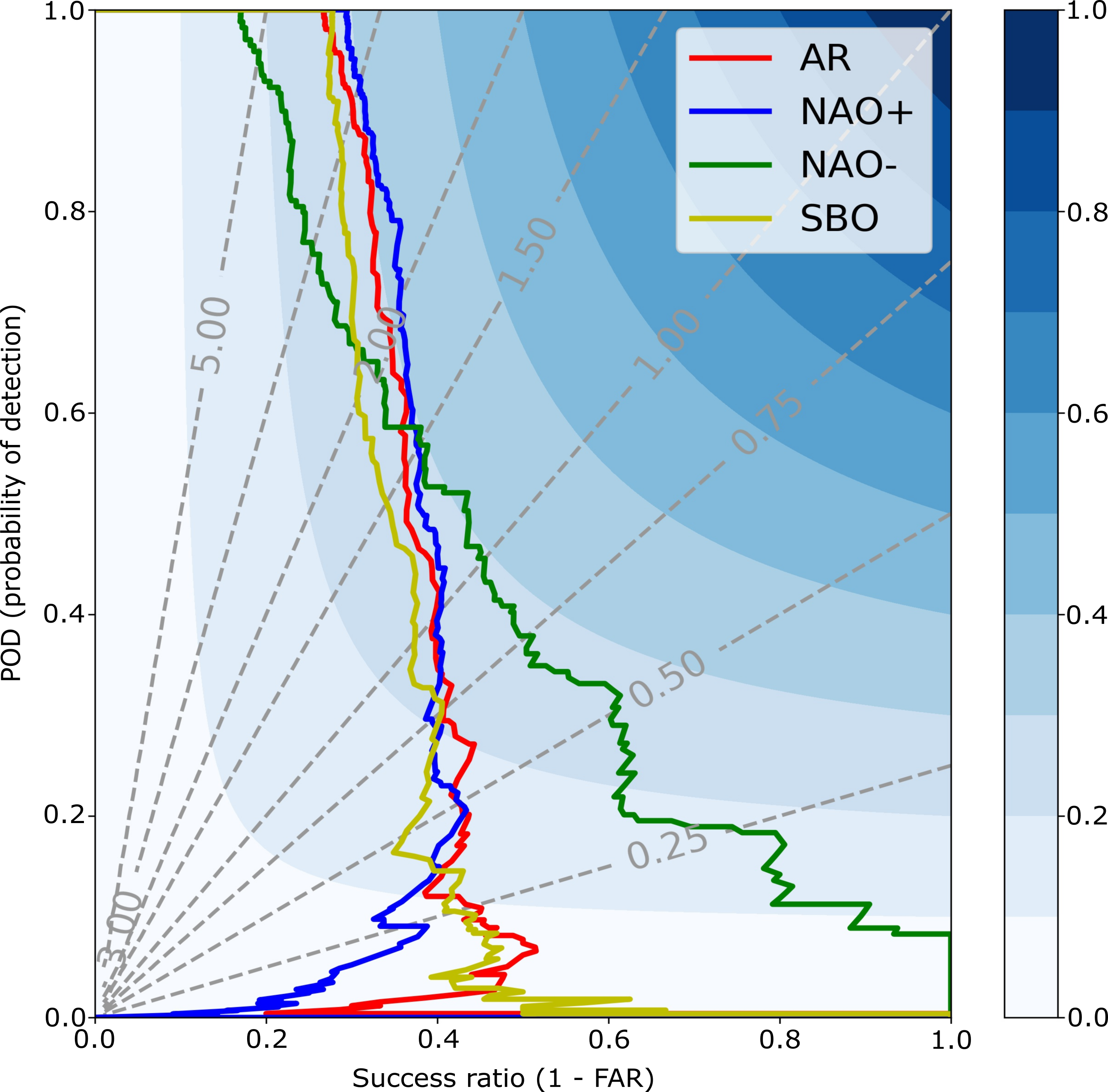}
    \caption{Performance diagram} \label{fig:roc_perf_di_b}
  \end{subfigure}
\caption{Receiver Operating Characteristics curve and performance diagram for deCNN when forecasting NAE weather regimes 10 days into the future. The dashed grey lines for Figure \ref{fig:roc_perf_di_a} denote a random model (random chance), which has an AUC of 0.5. The colorbar for Figure \ref{fig:roc_perf_di_a} refers to the Peirce score, defined as POD minus POFD. The dashed gray lines for Figure \ref{fig:roc_perf_di_b} denotes various levels of frequency bias. A frequency bias of 1.0 is considered an unbiased model. The colorbar for Figure \ref{fig:roc_perf_di_b} refers to CSI.} \label{fig:roc_perf_di}
\end{figure}

We also include a figure of classification accuracy and AUC-ROC curves in Figure \ref{fig:auc_acc}. Here, one can clearly see the relatively strong performance of deCNN at longer lead times compared to all benchmarks. This is true for accuracy in Figure \ref{fig:auc_acc_b} and even more evident for AUC-ROC in Figure \ref{fig:auc_acc_a}. Looking at accuracy in Figure \ref{fig:auc_acc_b}, we can also clearly see all models converging towards weekly climatology at longer lead times, with CNN only narrowly outperforming it at the 13-14 days lead time.

Next, we visualize ROC curves and a performance diagram for all classes at the ten days lead time for our deCNN in Figure \ref{fig:roc_perf_di}. First, we observe that the ROC curves lie well above the diagonal line for all classes at the various thresholds. We also notice the model has the highest skill for NAO-, as the green (NAO-) ROC curve lies considerably above the remaining ROC curves. This is mostly expected given NAO- is considered simpler to forecast than for example SB and AR, possibly due to being twice as persistent as other regimes \cite{Ferranti2018, dawson2012simulating}. In the performance diagram, NAO- also stands out compared to the other three regimes. Interestingly, our model turns out to be relatively good at forecasting AR, which contrasts with NWP models generally showing less forecasting skill at this regime compared to, for example, NAO+ \cite{Ferranti2018}. On the other hand, our model generally has a hard time forecasting SB consistent with NWP models \cite{Ferranti2018}.

\subsection*{Interpretation results}
We will use an interpretation technique called Integrated Gradients \cite{sundararajan2017axiomatic}, considered a state-of-the-art interpretation technique for deep neural networks \cite{metnet2}. This technique allows us to quantify the importance of each predictor on a grid-point basis for any given prediction. This is done by computing the path integral of gradients at all points along a straightline path from a baseline input $x^{'}$, typically an image with all zeros, to the actual input $x$. The formula \cite{sundararajan2017axiomatic} for any given $x$ is 

\begin{equation}
\text {IntegratedGrads}(x)::=\left(x-x^{\prime}\right) \times \int_{\alpha=0}^{1} \frac{\partial F\left(x^{\prime}+\alpha \times\left(x-x^{\prime}\right)\right)}{\partial x} d \alpha.
\end{equation}

Pixel attribution techniques, such as Integrated Gradients (IG), are typically prone to noise which can be addressed with a simple modification called Smooth-Grad (SG) \cite{smilkov2017smoothgrad}. Here, one injects Gaussian noise to the input several times before calculating the attributions of each perturbed sample. Next, the mean is computed across all sampled attributions. This approximates smoothing with a Gaussian kernel and effectively removes visual noise from the attribution maps making them more readily interpretable \cite{smilkov2017smoothgrad}. Finally, it has been shown that squaring the attributions before averaging them, denoted Smooth-Grad-Squared (SG-SQ), is a considerable improvement over SG \cite{hooker2019benchmark}. As a result, we use the SG-SQ approach in this work.

\begin{figure}[!t]
\footnotesize
  \hspace{0.25\textwidth}
  \begin{subfigure}{0.5\textwidth}
   \centering
    \includegraphics[width=\linewidth]{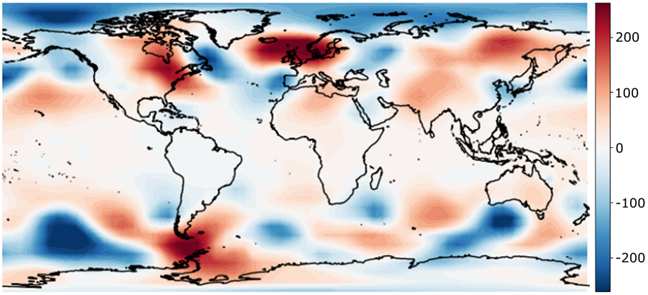}
    \caption{Geopotential height anomaly on the 2013-12-18} \label{fig:sbo_to_naom_a}
  \end{subfigure}%
   \vfill
  \begin{subfigure}{0.5\textwidth}
    \includegraphics[width=\linewidth]{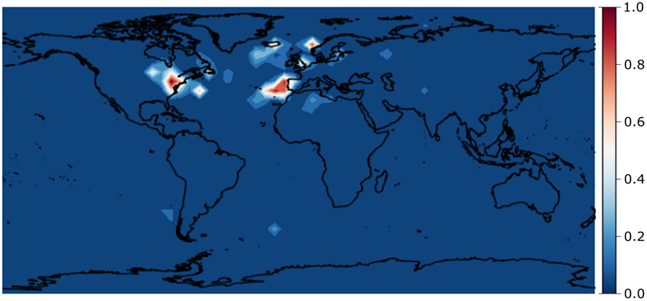}
    \caption{Feature importance deCNN} \label{fig:sbo_to_naom_b}
  \end{subfigure}%
  \hspace*{\fill}   
  \begin{subfigure}{0.5\textwidth}
    \includegraphics[width=\linewidth]{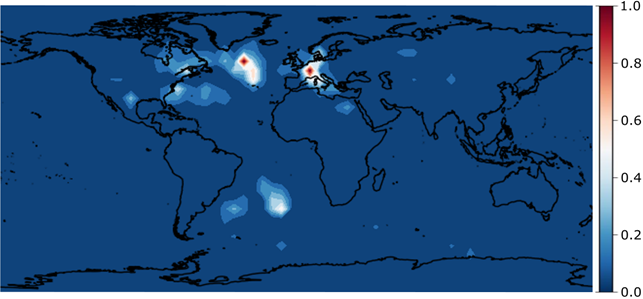}
    \caption{Feature importance deCNN (no pretrain)} \label{fig:sbo_to_naom_c}
  \end{subfigure}
\caption{SB transition to NAO- occurring from the 2018-11-13 to the 2018-11-24. The top Figure \ref{fig:sbo_to_naom_a} denotes the geopotential height anomaly input used by our model to make a regime prediction 11 days prior. The lower left Figure \ref{fig:sbo_to_naom_b} denotes the relevant regions for making that prediction for the deCNN model, and lower right Figure \ref{fig:sbo_to_naom_c} denotes the relevant regions for the deCNN model without transfer learning. Feature importances are normalized to lie between 0 and 1. Figures are created using cartopy 0.19.0 and matlotlib 3.3.4.} \label{fig:sbo_to_naom}
\vspace{-2mm}
\end{figure}

\begin{figure}[!htb]
  \hspace{0.25\textwidth}
  \begin{subfigure}{0.5\textwidth}
   \centering
    \includegraphics[width=\linewidth]{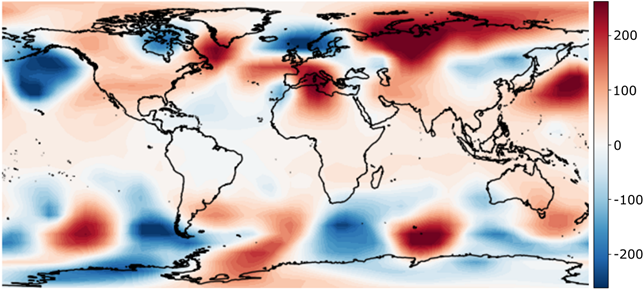}
    \caption{Geopotential height anomaly on the 2013-12-18} \label{fig:naop_to_naop_a}
  \end{subfigure}%
   \vfill
  \begin{subfigure}{0.5\textwidth}
    \includegraphics[width=\linewidth]{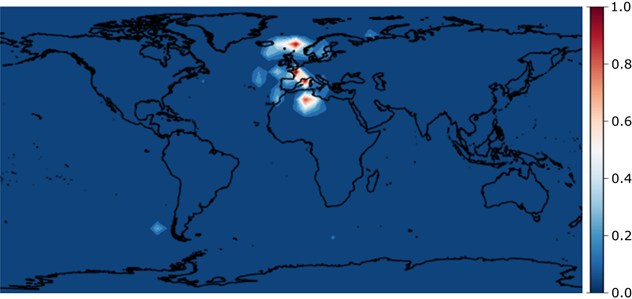}
    \caption{Feature importance deCNN} \label{fig:naop_to_naop_b}
  \end{subfigure}%
  \hspace*{\fill}   
  \begin{subfigure}{0.5\textwidth}
    \includegraphics[width=\linewidth]{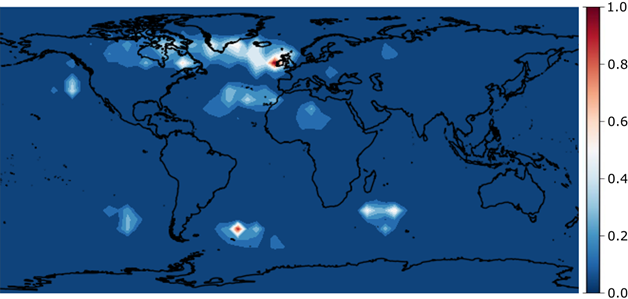}
    \caption{Feature importance deCNN (no pretrain)} \label{fig:naop_to_naop_c}
  \end{subfigure}
\caption{Persistence (i.e., unchanged) transition of NAO+ from the 2013-12-18 to the 2013-12-29. The top Figure \ref{fig:naop_to_naop_a} denotes the geopotential height anomaly input used by our model to make a regime prediction 11 days prior. The lower left Figure \ref{fig:naop_to_naop_b} denotes the relevant regions for making that prediction for the deCNN model, and lower right Figure \ref{fig:naop_to_naop_c} denotes the relevant regions for the deCNN model without transfer learning. Feature importances are normalized to lie between 0 and 1. Figures are created using cartopy 0.19.0 and matlotlib 3.3.4.} \label{fig:naop_to_naop}
\end{figure}

\begin{figure}[!htb]
  \hspace{0.25\textwidth}
  \begin{subfigure}{0.5\textwidth}
   \centering
    \includegraphics[width=\linewidth]{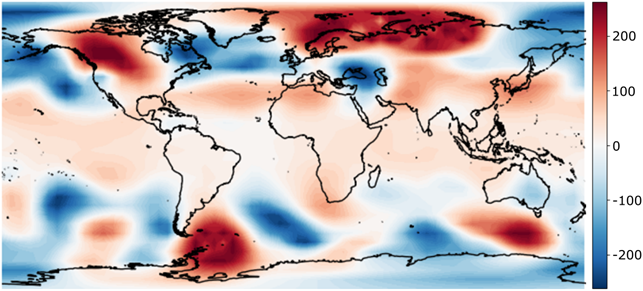}
    \caption{Geopotential height anomaly on the 2015-11-13} \label{fig:naop_to_ar_a}
  \end{subfigure}%
   \vfill
  \begin{subfigure}{0.5\textwidth}
    \includegraphics[width=\linewidth]{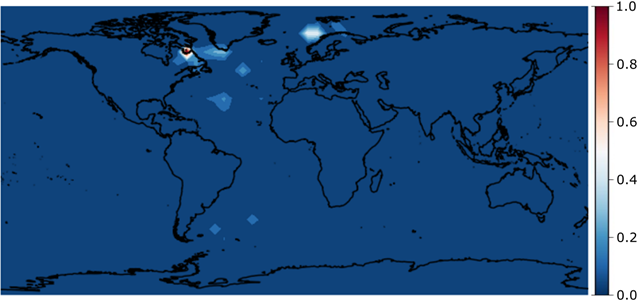}
    \caption{Feature importance deCNN} \label{fig:naop_to_ar_b}
  \end{subfigure}%
  \hspace*{\fill}   
  \begin{subfigure}{0.5\textwidth}
    \includegraphics[width=\linewidth]{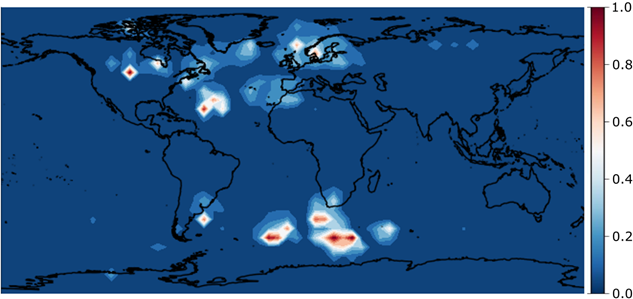}
    \caption{Feature importance deCNN (no pretrain)} \label{fig:naop_to_ar_c}
  \end{subfigure}
\caption{NAO+ transition to AR from the 2015-11-13 to the 2015-11-24. The top Figure \ref{fig:naop_to_ar_a} denotes the geopotential height anomaly input used by our model to make a regime prediction 11 days prior. The lower left Figure \ref{fig:naop_to_ar_b} denotes the relevant regions for making that prediction for the deCNN model, and lower right Figure \ref{fig:naop_to_ar_c} denotes the relevant regions for the deCNN model without transfer learning. Feature importances are normalized to lie between 0 and 1. Figures are created using cartopy 0.19.0 and matlotlib 3.3.4.} \label{fig:naop_to_ar}
\end{figure}

One way to increase the robustness and trustworthiness of our model is to investigate integrated gradients for some of the most probable NAE weather regime transitions. Here, the stated frequencies of the regime transitions under investigation will be based on extensive previous research \cite{Ferranti2018}. Furthermore, we have chosen to focus specifically on the Z500 variable because of its importance in the meteorological community \cite{Rasp2020} and also since it is the variable used to derive the NAE regimes, but we could also have included other variables of interest. We also include the integrated gradients map for our deCNN (no pretrain) model to further emphasize the importance of transfer learning.

\textbf{SB to NAO-}. The first distinct transition we investigate is from SB to NAO-, happening around 30\% of the time at the 11-days lead time \cite{Ferranti2018}. Figure \ref{fig:sbo_to_naom} shows the raw input and the two integrated gradient maps for an example of the SB to NAO- transition for deCNN and deCNN (no pretrain). The deCNN model accurately predicted this transition with a probability of 32.5\%. First, looking at Figures \ref{fig:sbo_to_naom_a} and \ref{fig:sbo_to_naom_b}, we note a region of interest at the negative anomaly cluster occurring west of Spain and Portugal, a distinctive and unique feature for NAO- compared to other clusters. Second, we observe a positive anomaly stretching from Scandinavia towards Iceland, which also contains multiple but smaller attributions at various locations along this stretch. This pattern aligns with the positive anomaly area over Scandinavia typical for SB, which then transitions toward Greenland, a distinct pattern for NAO-. Finally, we note a region of interest at the east coast of the United States overlapping with a positive anomaly. This pattern makes sense from a meteorological point of view, as geopotential height anomalies at the east coast of the US have a large influence on enhanced or reduced westerly flows and, by extension, the dominating EOFs in the NAE region. If we look deeper into the integrated gradients map for the deCNN (no pretrain) model in Figure \ref{fig:sbo_to_naom_c}, we notice the patterns are not as straightforward to interpret. For example, the model emphasizes the region around France as important, which is not a unique characteristic of the SB to NAO- transition. Additionally, the highlighted regions are less outspoken compared to the deCNN model. On the other hand, the model also emphasizes more global connections, such as two regions in the South Atlantic Ocean. As a result, one might argue that while transfer learning does improve forecasting accuracy, it also increases the emphasis on local teleconnections, potentially making the methodology less useful if the objective is to find novel teleconnections rather than maximizing forecasting accuracy.

\textbf{NAO+ persistence}. Next, we investigate a typical NAO+ persistence transition, i.e., NAO+ being the most probable precursor for NAO+ up until 11 days into the future. This happens around 37.7\% of the time \cite{Ferranti2018}. Our deCNN model also correctly predicts this transition with a probability of 39.1\%. We visualize the integrated gradients and raw input in Figure \ref{fig:naop_to_naop}. First, we note that the attribution map in Figure \ref{fig:naop_to_naop_b} match the exact patterns that distinguish NAO+ from the remaining regimes, i.e., a negative anomaly in the northernmost area of the Atlantic Ocean and a positive anomaly stretching from mainland central Europe to the Atlantic Ocean. Interestingly, the ocean area halfway between Europe and the US, consisting of a positive anomaly, is not being attributed as a region of interest by our model despite being a feature of the NAO+ regime. If we compare NAO+ with the AR regime, we can potentially see why. A positive anomaly clearly identifies the AR regime in exactly this area, thus overlapping with NAO+, and so the model cannot, and does not, use this positive anomaly to differentiate between the two regimes. Looking closer at Figure \ref{fig:naop_to_naop_c}, we observe that the deCNN (no pretrain) model actually emphasizes this positive anomaly as being important. This is an interesting finding, as it provides some visual evidence for the importance of transfer learning. On the other hand, we also observe a general pattern similar to the previous transition in Figure \ref{fig:sbo_to_naom}, which shows a substantial increase in the number of relevant global regions and a decrease in feature importance values.

\textbf{NAO+ to AR}. The final transition we investigate is the less likely transition of NAO+ to AR, occurring around 23.8\% of the time up to 11 days in the future. Our deCNN model incorrectly predicts this transition and instead predicts the transition to be NAO+ to SB with a probability of 26.4\%. However, despite the incorrect prediction, NAO+ to SB is actually considered the most likely transition compared to NAO+ to AR \cite{Ferranti2018}. If we look at the integrated gradients map in Figure \ref{fig:naop_to_ar_b}, we see a) a region of interest in Eastern Canada with a negative anomaly and b) a region of interest with a positive anomaly in the Norwegian Sea. This increases the pressure gradient across the Atlantic sea, eventually causing enhanced westerly flow \cite{lopez2011effects}, a key feature of the NAO+ regime. Despite correctly emphasizing the NAO+ patterns, the integrated gradients in Figure \ref{fig:naop_to_ar_b} seem less profound and more sparse than the previous integrated gradient figures. This could also explain the relative uncertainty of the deCNN model for this particular observation. We observe two interesting findings when comparing deCNN in Figure \ref{fig:naop_to_ar_b} with the deCNN (no pretrain) model in Figure \ref{fig:naop_to_ar_c}. First, similar to previous transitions, the model dramatically increases the number of relevant regions and their absolute values. On the other hand, the model seems to highly emphasize regions outside Europe, such as the ocean south of Africa. While it is theoretically possible that such regions could be relevant for predicting the NAO+ to AR transition, given the low accuracy of the deCNN (no pretrain) model, it seems more likely that this shows the model fails in identifying the relevant regions for this particular transition. 

\section*{Discussion}
This paper introduced a supervised learning strategy based on deformable convolutional neural networks to forecast large-scale circulation regimes in the NAE region 1 to 15 days into the future. Circulation regimes pose an interesting problem for data-driven methods due to numerous historical observations combined with a critical role in understanding the global climate system and clustering weather predictability. In this regard, our method is purely data-driven while having a global and general perspective, meaning it can easily be extended to forecast other global teleconnections or weather regimes. To achieve our stated goal and succumb to the data-driven paradigm, we applied several techniques for leveraging the capabilities and requirements of data-driven methods, including transfer learning on the large 20CRv3 reanalysis dataset spanning from 1836 to 1980 and the smaller but more recent ERA5 reanalysis dataset from 1980 to 2018. This proved to be of paramount importance to the strong performance of the deCNN model. Next, we applied a smart stratified sampling technique to divide our dataset into a training set and validation set based on several multidecadal oscillations and teleconnection patterns, such as the PDO and AMO. Finally, we trained an individual model for each 1 to 15 days lead time. However, to reduce the impact of a sub-optimal set of hyperparameters when training these models, we applied Bayesian Optimization to fast and efficiently find an optimal set of hyperparameters. 

The findings from this paper were twofold. First, we could extract and learn transformation-invariant spatial patterns across large geographical areas using deformable convolutions, which is not possible with regular CNNs. This culminated in the deCNN achieving considerably better performance than a regular CNN model, logistic regression, random forests, persistence, and climatology benchmarks. The importance of using deformable convolutions was especially evident at longer lead times. As a result, one might conclude that deCNNs are mostly relevant when considering forecasting at lead times beyond 5-6 days into the future, and regular CNNs work almost equally well at shorter lead times. We also investigated the various NAE regimes and their predictability in more detail. We found the NAO- regime to be the most predictable and SB the least predictable, consistent with NWP models and previous studies. On the other hand, we observed slightly better results for AR than NAO+, which is not the case for NWP models.

Second, using a sophisticated interpretation technique called integrated gradients, we could attribute each variable's contributions for a particular observation on a grid-point basis. This is especially important if we want to understand global climate processes better and explain drivers behind specific weather regimes that account for major uncertainty in NWP models days to weeks ahead. In this regard, we investigated some of the most likely regime transitions found in the literature. Here, we observed the deCNN model emphasizing unique and distinct features and regions for the Z500 variable for three different transitions. Besides successfully forecasting two of the three transitions, the integrated gradient maps were mostly consistent with the unique features that distinguish the various weather regimes and their transitions, thus enhancing trust in our model. On the other hand, the deCNN model without transfer learning showed an increase in the number of highlighted regions and put less emphasis on the unique characteristics of the dominant regime transitions. The increased number of relevant regions could indicate that transfer learning restricts our deCNN model to focus on more local teleconnection patterns to improve forecasting accuracy. As a result, if the aim of future related studies should center more around scientific discovery rather than maximizing forecasting accuracy, transfer learning should potentially be balanced against these two somewhat conflicting objectives.

While the results of our proposed framework are promising, we hypothesize that extending our analysis to investigate year-round NAE regimes or other weather regimes at the subseasonal time scale could be extremely interesting given the low forecasting skill at this time horizon \cite{Albers_2021}. In this regard, one specific application could be the comparison between our proposed methodology and the recent work by Büeler et al. (2021). \cite{bueler2021year}. Furthermore, recent research \cite{mamalakis2022investigating, mamalakis2021neural} has highlighted a few limitations of using integrated gradients and other interpretation techniques for deep neural networks. Therefore, future research should consider investigating a more holistic view using multiple interpretation techniques. Finally, focusing specifically on interpretation techniques as the primary output could potentially be useful for discovering novel teleconnections targeted towards other specific weather regimes, which might improve our understanding of climate dynamics globally.

\section*{Data availability}
The NOAA-CIRES-DOE Twientieth Century Reanalysis v3 dataset is publicly available at \url{https://psl.noaa.gov/data/gridded/data.20thC_ReanV3.html}. The ERA5 dataset is publicly available at the Copernicus Climate Data Store \url{https://cds.climate.copernicus.eu/}. ENSO time series is publicly available at \url{https://psl.noaa.gov/people/cathy.smith/best/}. PDO time series is publicly available at \url{https://www.ncdc.noaa.gov/teleconnections/pdo/}, and finally AMO can be found at \url{https://psl.noaa.gov/data/timeseries/AMO/}.

\bibliography{sample}

\section*{Acknowledgements}

We are thankful to Axel Wagner, Kim Bentzen, and Morten Hygum from Danske Commodities A/S for fruitful discussions related to this work. This work was funded by Danske Commodities A/S.

\section*{Author contributions statement}
A.N conceived the experiment(s). A.I. and H.K. supervised the research. A.N conducted the experiment(s). All authors reviewed the manuscript.

\section*{Additional information}

\textbf{Competing interests:} The authors declare no competing interests.

\end{document}